\pdfoutput=1

\documentclass[11pt]{article}

\usepackage[final]{acl}
\usepackage{tikz}
\usepackage{times}

\usepackage{soul}
\usepackage{tcolorbox} 
\usepackage[utf8]{inputenc} 

\usepackage[T1]{fontenc}

\usepackage[utf8]{inputenc}

\usepackage{microtype}

\usepackage{inconsolata}

\usepackage{graphicx}

\usepackage{booktabs}
\usepackage{pifont}
\usepackage{multirow}
\usepackage{caption}
\usepackage{hyperref}
\newlength\savewidth

\newcommand{\cmark}{\ding{51}}%
\newcommand{\xmark}{\ding{55}}%

\newtcolorbox{boxK}{
    sharpish corners, 
    boxrule = 0pt,
    toprule = 4.5pt, 
}

\title{Is C4 Dataset Optimal for Pruning? \\An Investigation of Calibration Data for LLM Pruning}

\author{
\small{Abhinav Bandari$^{1, 3}$\thanks{Work done while the author was at University of Washington. Correspondence to: Abhinav Bandari \\ <abhinavbandari@utexas.edu>}, Lu Yin$^{2}$}, 
\small{\textbf{Cheng-Yu Hsieh$^1$, Ajay Kumar Jaiswal$^{3}$}}\\
\small{\textbf{Tianlong Chen$^{4}$, Li Shen$^{5}$, Ranjay Krishna$^{1}$, Shiwei Liu$^{6}$}} 
\\
\small{$^1\,$University of Washington\hspace{0.2cm} $^2\,$University of Surrey\hspace{0.2cm} $^3\,$The University of Texas at Austin}\\[0.1ex]
\small{$^4\,$The University of North Carolina at Chapel Hill\hspace{0.2cm} $^5\,$Sun Yat-sen University\hspace{0.2cm} $^6\,$University of Oxford}
}

\begin{document}
\maketitle

\begin{abstract}
Network pruning has emerged as a potential solution to make LLMs cheaper to deploy. However, existing LLM pruning approaches universally rely on the C4 dataset as the calibration data for calculating pruning scores, leaving its optimality unexplored. 
In this study, we evaluate the choice of calibration data on LLM pruning, across a wide range of datasets that are most commonly used in LLM training and evaluation, including four pre-training datasets as well as three categories of downstream tasks encompassing nine datasets. Each downstream dataset is prompted with In-Context Learning (ICL) and Chain-of-Thought (CoT), respectively. Besides the already intriguing observation that the choice of calibration data significantly impacts the performance of pruned LLMs, our results also uncover several subtle and often unexpected findings, summarized as follows: (1) C4 is not the optimal choice for LLM pruning, even among commonly used pre-training datasets; (2) arithmetic datasets—when used as calibration data—performs on par or even better than pre-training datasets; (3) pruning with downstream datasets does not necessarily help the corresponding downstream task, compared to pre-training data; (4) ICL is widely beneficial to all data categories, whereas CoT is only useful on certain tasks.
Our findings shed light on the importance of carefully selecting calibration data for LLM pruning and pave the way for more efficient deployment of these powerful models in real-world applications. We release our code at: \url{https://github.com/abx393/llm-pruning-calibration-data}.


\end{abstract}

\section{Introduction}

In the 2020s, the landscape of AI has transitioned into a new era, propelled forward by the advancements made in large language models (LLMs)~\cite{brown2020language, team2023gemini, touvron2023llama2}. The astonishing language capacities of LLMs have significantly shaped the solutions to various real-life tasks such as natural language understanding~\cite{brown2020language,touvron2023llama}, text generation \cite{kocon2023chatgpt,anil2023palm}, vision tasks \cite{radford2021learning,zhou2022conditional}, coding \cite{chen2022transferability}, and math \cite{romera2024mathematical}. However, the enormous size of these powerful LLMs poses a significant challenge for deployment in many real-world applications. For instance, deploying a 7B LLM requires around 10GB of main memory (DRAM) even after adopting INT8 quantization, which unfortunately exceeds the memory capacity of most commodity edge devices.

Network pruning, as one of the most well-established approaches in model compression, demonstrated the possibility of removing around 50\% of the parameters \cite{frantar2023massive,sun2023simple,zhang2023dynamic}, or even more~\cite{yin2023outlier,agarwalla2024enabling} with minimal performance degradation. Interestingly, while consistently producing robust performance in small-scale deep neural networks \cite{han2015learning,frankle2018lottery,mocanu2018scalable,gale2019state}, magnitude pruning \cite{han2015learning} seems to lose importance in the context of LLM pruning. All state-of-the-art LLM pruning approaches unanimously choose to use a small set of data (known as \textit{calibration data}) from the C4 training dataset \cite{raffel2020exploring} to calculate their pruning scores \cite{frantar2023massive,sun2023simple,yin2023outlier}.

Using C4 as the calibration data for pruning makes sense if the models are pre-trained on it to preserve better the desired distribution learned during pre-training. However, not all large language models are pre-trained with the C4 dataset, raising the question of whether the C4 is the optimal choice for the calibration data for LLM pruning.
In addition, it is well-known that LLMs are very sensitive to how the input is structured and provided to them~\cite{zhou2022large,shi2023large}. As a result, it is unclear how the input format of calibration data would affect LLM pruning.

To answer these questions, in this work, we conduct a comprehensive study to investigate the effect of calibration data on LLM pruning across a broad range of evaluation tasks, along two dimensions of interest: varying types of datasets and different data input formats. Specifically, we investigate the following possible alternatives of calibration data for LLM pruning, as illustrated in Figure \ref{fig:overview}: 

\begin{figure*}[h]
    \centering
    \includegraphics[scale=0.6]{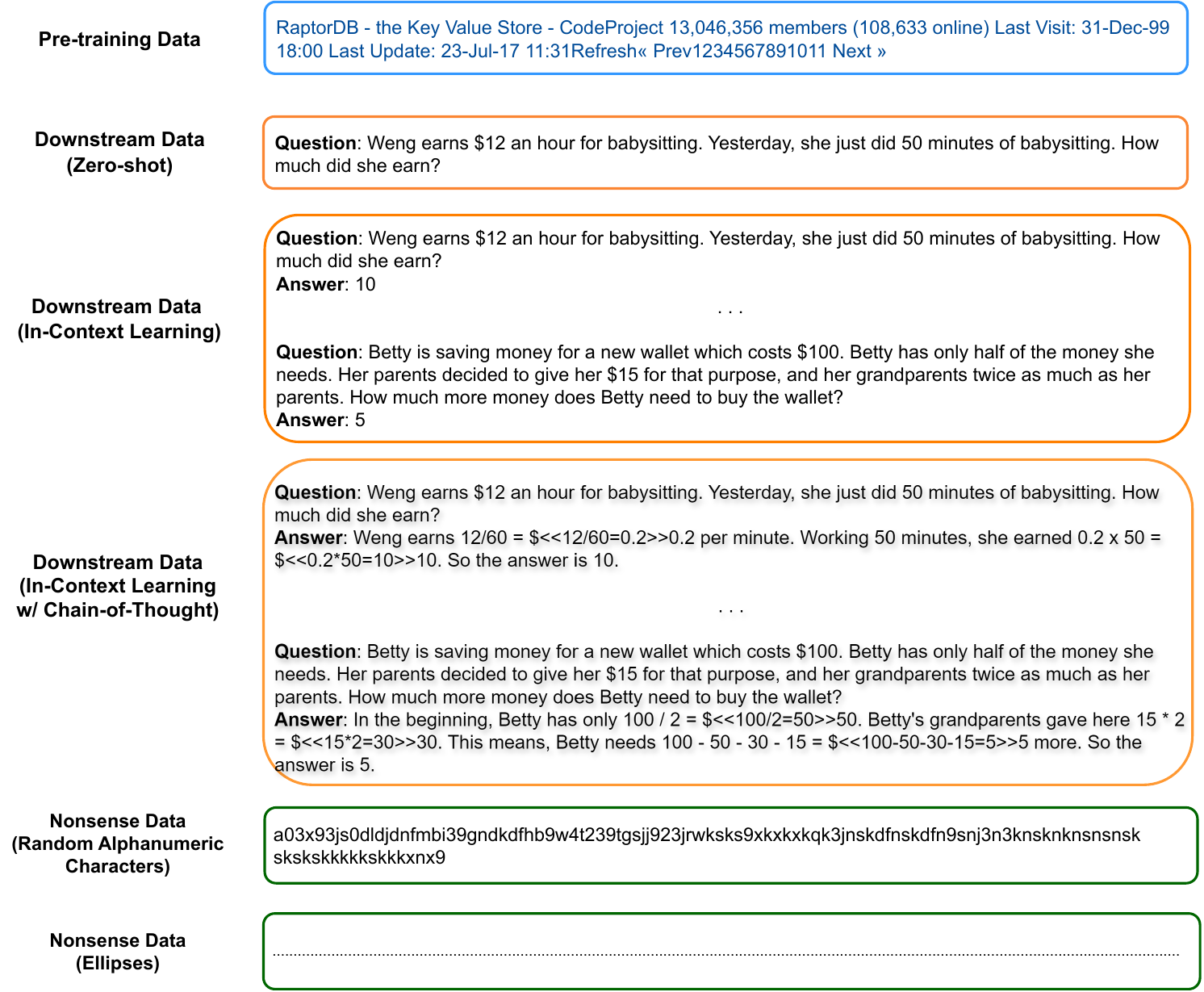}
    \caption{Examples of various calibration data formats examined in this paper.}
    \label{fig:overview}
\end{figure*}

\begin{itemize}
    \item \textbf{Pre-training Data}: Apart from the C4 dataset, several other datasets are widely used for pre-training LLMs. We examine three of the most representative datasets: Pile \cite{gao2020pile}, OSCAR \cite{suarez2020monolingual}, and RedPajama \cite{together2023RedPajama}.

    \item \textbf{Downstream Data}: While pruning with pre-training datasets is intuitively preferred to preserve pre-training knowledge, it is essential to empirically verify this assumption and identify whether pruning with any downstream datasets may yield superior outcomes for LLM pruning. To investigate this, we consider three categories of downstream tasks, encompassing a total of nine datasets (see Section \ref{sec:setup} for details). An intriguing research question arises: \textit{will pruning with downstream data produce a better sparse model for the corresponding downstream task than pruning with pre-training data?}
    


    \item \textbf{Prompted Downstream Data}: Acknowledging the significant impact of prompts on LLM performance, we explore two variants of prompting strategies to construct different formats of calibration data: In-Context Learning (ICL) \cite{brown2020language} and In-Context Learning w/ Chain-of-Thought (ICL w/ CoT) \cite{wei2022chain}.

    \item \textbf{Nonsense Data}:  In addition, we explore two variants of nonsensical calibration data—ellipses and random alphanumeric strings—to investigate the necessity of semantically meaningful calibration data for effective LLM pruning.

\end{itemize}

To investigate the impact of these datasets, we prune LLMs using various calibration datasets and evaluate the resulting sparse models across nine downstream tasks. Our key and encouraging finding is that, while C4 consistently produces robust sparse models, it is not the best calibration dataset for pruning. 
In addition, our study unveils several more subtle and unexpected findings, which can be summarized as follows:



%


\begin{itemize}
\item C4, although consistent in producing robust sparse models, is not the optimal choice for LLM pruning, and it is also not the best among various pre-training datasets. Pile consistently outperforms C4 with higher average accuracy. 
    
\item Certain types of downstream data lead to better sparse LLMs than others.  Arithmetic downstream datasets in general perform on par or even better than pre-training datasets in this context of LLM pruning.

\item
Pruning with downstream data does not necessarily lead to the best performance on that downstream task than pruning with a pre-training dataset like Pile.

\item
ICL calibration data broadly benefits all data categories, while ICL w/ CoT calibration data is only advantageous for arithmetic reasoning datasets.

\end{itemize}



\section{Related Work}

\subsection{Large Language Model Pruning}
Network pruning is a widely utilized technique to reduce model size with negligible performance loss \citep{mozer1989skeletonization,han2015learning,molchanov2016pruning}. While numerous pruning approaches have been proposed, the success of pruning is inextricably linked to sufficient re-training \cite{liu2022unreasonable,wang2023state}. However, training large language models is prohibitively expensive and not feasible for most practitioners. Fortunately, recent research efforts have proposed effective methods that enable accurate pruning of LLMs without the need for extensive fine-tuning.
SparseGPT \citep{frantar2023massive} employs second-order pruning followed by column-wise weight updates, allowing the removal of 50\% of weights while maintaining the original perplexity. Wanda \citep{sun2023simple}, motivated by the goal of preserving crucial outliers in LLMs, proposes pruning weights based on the multiplication of weight magnitude with their input activation, demonstrating strong performance. OWL \citep{yin2023outlier} introduces a novel non-uniform layerwise sparsity approach for LLM pruning, showing promising results at high levels of sparsity.
In addition to exploring accurate pruning methods, other studies focus on efficiently fine-tuning sparse LLMs to further enhance their performance \cite{zhang2023dynamic,zimmer2023perp}. In contrast to these previous works, our paper investigates the efficacy of input data for LLM pruning. This novel perspective is crucial for understanding and improving LLM pruning methodologies, as LLMs are sensitive to their input~\citep{zhao2021calibrate}.
 
\subsection{Prompting for Sparse LLMs}

Prompting involves providing instructions to a pre-trained language model, either as a single instruction (zero-shot) or through one or more examples (one/few-shot) that demonstrate the task. \citet{brown2020language} demonstrated that prompt design is highly effective for guiding a non-modifiable GPT-3 model in zero, one, and few-shot settings. Initially, efforts in prompt-tuning focused on the discrete selection of prompt template tokens, as explored by \citet{jiang2020can}. Later studies, such as those by \citet{lester2021power}, shifted towards using continuous prompts that were refined through backpropagation.


\citet{xu2023compress} first discovered that the generation quality of a compressed LLM could be significantly improved by adding carefully designed hard prompts and proposed a soft prompt learning method to improve the compressed LLM. \citet{hoang2023dynamic} argued that the performance drop caused by pruning is because the pre-trained knowledge is displaced rather than being forgotten. \citet{williams2023does} examined the impact of multiple pre-training data sources on pruning. However, their study was confined to pre-training data sources. Our research extends this investigation by not only analyzing four commonly used pre-training datasets but also exploring various downstream datasets with In-Context Learning and Chain of Thought prompts, leading to more intriguing findings and a deeper understanding of the effects of different data sources on pruning.

\begin{table}[h]
\centering
\caption{Pruning metrics of Wanda and SparseGPT.}
\vspace{-1em}
\renewcommand{\arraystretch}{1.15}
\setlength{\tabcolsep}{4.5pt}
\resizebox{\linewidth}{!}{
\begin{tabular}{l c c}
        & & \\
    \toprule
    Method  & Weight Update & Pruning Metric $\mathbf{S}_{ij}$ \\
    \hline
    SparseGPT & \cmark & $\left[|\mathbf{W}|^{2} / \mathrm{diag}\bigl[(\mathbf{X}\mathbf{X}^{T}+\lambda \mathbf{I})^{-1}\bigr]\right]_{ij}$ \\
    Wanda  & \xmark & $|\mathbf{W}_{ij}| \cdot \|\mathbf{X}_{j}\|_{2}$ \\
    \bottomrule
\end{tabular}
}
\label{tab:prune}
\end{table}

\section{Methodology}

In this section, we describe in detail how we assess the effects of various calibration datasets and data formats on LLM pruning. 

\subsection{Pruning Methods} 

We choose the two most widely-used pruning methods, i.e., Wanda \cite{sun2023simple} and SparseGPT \cite{frantar2023sparsegpt} as our pruning methods. Both pruning methods necessitate a small subset of calibration data to calculate pruning scores, which are shown in Table~\ref{tab:prune}. In this context, $\mathbf{X}$ symbolizes layer activations and $\mathbf{W}$ represents weights. The expression $\mathbf{X}^{T}\mathbf{X}+\lambda \mathbf{I}$ in the denominator forms the Hessian $\mathbf{H}$, essential for the layer-wise reconstruction issue, with $\lambda$ serving as a dampening factor to prevent computational collapse during inversion. 
Wanda augments the standard weight magnitude pruning metric by integrating input activations, whereas SparseGPT incorporates an additional weight update step within its column-wise pruning process. 
The weights with the lowest scores will be pruned, resulting in a sparse LLM. 




\begin{table*}[htbp]
\centering
\resizebox{1\textwidth}{!}{%
\begin{tabular}{l|c|cccc|cccc}
\toprule
\textbf{Evaluation} & \textbf{Dense} & \multicolumn{4}{c|}{\textbf{Wanda  \emph{w.} Calibration Data}} & \multicolumn{4}{c}{\textbf{SparseGPT \emph{w.} Calibration Data}} \\
& \textbf{} & \textbf{C4} & \textbf{RedPajama} & \textbf{Oscar} & \textbf{Pile} & \textbf{C4} & \textbf{RedPajama} & \textbf{Oscar} & \textbf{Pile} \\
\midrule
GSM8K        & 0.0576 & \textbf{0.0457} $\pm$ 0.0008 & 0.0412 $\pm$ 0.0062 & 0.0450 $\pm$ 0.0088 & 0.0404 $\pm$ 0.0048 & \textbf{0.0440 $\pm$ 0.0052} & 0.0430 $\pm$ 0.0048 & 0.0412 $\pm$ 0.0046 & 0.0384 $\pm$ 0.0038 \\
SVAMP        & 0.3867 & 0.2756 $\pm$ 0.0102 & 0.2733 $\pm$ 0.0133 & \textbf{0.2922} $\pm$ 0.0102 & 0.2878 $\pm$ 0.0038 & 0.3011 $\pm$ 0.0193 & 0.3033 $\pm$ 0.0370 & 0.3089 $\pm$ 0.0278 & \textbf{0.3445 $\pm$ 0.0139} \\
MAWPS        & 0.4462 & 0.3160 $\pm$ 0.0293 & 0.3154 $\pm$ 0.0308 & 0.3436 $\pm$ 0.0097 & \textbf{0.3635} $\pm$ 0.0115 & 0.3295 $\pm$ 0.0235 & 0.3487 $\pm$ 0.0289 & 0.3500 $\pm$ 0.0416 & \textbf{0.3820 $\pm$ 0.0262} \\
\midrule
e-SNLI       & 0.6050 & 0.4934 $\pm$ 0.0096 & 0.4940 $\pm$ 0.0377 & 0.4812 $\pm$ 0.0205 & \textbf{0.5376} $\pm$ 0.0023 & 0.5447 $\pm$ 0.0326 & \textbf{0.5641 $\pm$ 0.0295} & 0.5485 $\pm$ 0.0487 & 0.5498 $\pm$ 0.0289 \\
ANLI R1      & 0.3900 & 0.3250 $\pm$ 0.0156 & 0.3240 $\pm$ 0.0255 & 0.3203 $\pm$ 0.0042 & \textbf{0.3420} $\pm$ 0.0087 & 0.3580 $\pm$ 0.0356 & \textbf{0.3640 $\pm$ 0.0183} & 0.3463 $\pm$ 0.0261 & 0.3380 $\pm$ 0.0020 \\
ANLI R3      & 0.4192 & 0.3361 $\pm$ 0.0106 & 0.3405 $\pm$ 0.0058 & 0.3220 $\pm$ 0.0042 & \textbf{0.3597} $\pm$ 0.0145 & \textbf{0.3575 $\pm$ 0.0153} & 0.3478 $\pm$ 0.0226 & 0.3480 $\pm$ 0.0136 & 0.3408 $\pm$ 0.0043 \\
\midrule
CSQA & 0.6208 & 0.5171 $\pm$ 0.0024 & 0.5184 $\pm$ 0.0184 & 0.5225 $\pm$ 0.0043 & \textbf{0.5239} $\pm$ 0.0078 & 0.5266 $\pm$ 0.0314 & \textbf{0.5304 $\pm$ 0.0204} & 0.5233 $\pm$ 0.0141 & 0.5258 $\pm$ 0.0331 \\
RACE         & 0.6501 & 0.4686 $\pm$ 0.0052 & 0.4386 $\pm$ 0.0109 & 0.4632 $\pm$ 0.0224 & \textbf{0.4692} $\pm$ 0.0079 & 0.5305 $\pm$ 0.0272 & \textbf{0.5407 $\pm$ 0.0101} & 0.5374 $\pm$ 0.0279 & 0.5376 $\pm$ 0.0215 \\
WinoGrande   & 0.5122 & \textbf{0.5141} $\pm$ 0.0094 & \textbf{0.5141} $\pm$ 0.0067 & \textbf{0.5141} $\pm$ 0.0087 & 0.5125 $\pm$ 0.0051 & 0.5183 $\pm$ 0.0143 & 0.5193 $\pm$ 0.0016 & \textbf{0.5240 $\pm$ 0.0311} & 0.5164 $\pm$ 0.0194 \\
\midrule
Average      & 0.4542 & 0.3657 $\pm$ 0.0041 & 0.3622 $\pm$ 0.0005 & 0.3671 $\pm$ 0.0055 & \textbf{0.3819} $\pm$ 0.0029  & 0.3900 $\pm$ 0.0063 & 0.3957 $\pm$ 0.0090 & 0.3920 $\pm$ 0.0153 & \textbf{0.3970 $\pm$ 0.0045} \\
\bottomrule
\end{tabular}}
\caption{Accuracy of Llama 2-Chat 7B model pruned with Wanda and SparseGPT to 50\% unstructured sparsity using different pre-training datasets, averaged over three random seeds. The value after $\pm$ indicates 2 standard deviations. Results for both pruning methods are shown alongside the original dense model for comparison. The best performance on each evaluation task for each pruning algorithm is bold.}
\label{tab:combined_pretraining}
\end{table*}

\begin{table}[h]
\centering
\resizebox{\columnwidth}{!}{%
\begin{tabular}{l|c|cccc}
\toprule
\textbf{Evaluation task} & \textbf{Dense Model} & \multicolumn{4}{c}{\textbf{Wanda  \emph{w.} Calibration Data}} \\
& \textbf{} & \textbf{C4} & \textbf{RedPajama} & \textbf{Oscar} & \textbf{Pile} \\
\midrule
GSM8K        & 0.0576 & \textbf{0.0269} & 0.0186 & 0.0208 & 0.0239 \\
SVAMP        & 0.3867 & \textbf{0.0200} &  0.0133 & \textbf{0.0200} & 0.0133 \\
MAWPS        & 0.4462 & \textbf{0.0019} & 0.0000 & 0.0000 & 0.0000 \\
\midrule
e-SNLI       & 0.6050 & 0.1313 &  0.2432 & 0.0687 & \textbf{0.3249} \\
ANLI R1      & 0.3900 & 0.0000 & 0.0000 & 0.0000 & \textbf{0.1190} \\
ANLI R3      & 0.4192 & 0.0000 & 0.0000 & 0.0000 & \textbf{0.0925} \\
\midrule
CSQA & 0.6208 & 0.2138 & 0.2072 & 0.2113 &  \textbf{0.2170} \\
RACE         & 0.6501 & 0.2528 & 0.2197 & 0.2514 & \textbf{0.2540} \\
WinoGrande   & 0.5102 & \textbf{0.5012} & 0.4925 & 0.4743 & 0.4972 \\
\midrule
Average      & 0.4542 & 0.1275 & 0.1327 & 0.1163 & \textbf{0.1713} \\
\bottomrule
\end{tabular}}
\caption{Accuracy of Llama 2-Chat 7B model pruned with Wanda to 70\% unstructured sparsity using different pre-training datasets. Results are shown alongside the original dense model for comparison. The best performance on each evaluation task is bold.}
\label{tab:combined_pretraining_70}
\end{table}

\label{sec:setup}
\subsection{Model, Dataset, and Evaluation}
\textbf{Model.} We use the  common models used in previous work \cite{sun2023simple,yin2023dynamic}, i.e., Llama 2-Chat 7B \citep{touvron2023llama2} and Llama 7B \citep{touvron2023llama} as the base models for pruning. 

\noindent \textbf{Dataset.} The source of our calibration data is divided into two categories: pre-training datasets and downstream datasets. For pre-training data, we selected four widely-used datasets: C4 \cite{raffel2020exploring}, Pile \cite{gao2020pile}, OSCAR \cite{suarez2020monolingual}, and RedPajama \cite{together2023RedPajama}. To ensure the diversity of the downstream calibration data, we focused on three major tasks: arithmetic reasoning, natural language inference, and commonsense reasoning, selecting three datasets for each category. 

\noindent For arithmetic reasoning, we chose the following three datasets:
\begin{itemize}
    \item \textbf{GSM8K} \citep{cobbe2021gsm8k} is a dataset of grade school math word problems, where each problem takes between 2 and 8 steps to solve.
    \item \textbf{SVAMP} \citep{patel2021svamp} is another dataset of grade school math word problems, where each problem requires no more than 2 arithmetic operations to solve.
    \item \textbf{MAWPS} \citep{koncel2016mawps} is another dataset of grade school math word problems of varying complexity.
\end{itemize}
For natural language inference, we use the following datasets:
\begin{itemize}
    \item \textbf{e-SNLI} \citep{camburu2018esnli} is a dataset of entailment relations along with human-annotated natural language explanations of the labels.
    \item \textbf{ANLI} \citep{nie2020anli} is a dataset of entailment relations that was iteratively and adversarially generated with a human-and-model-in-the-loop procedure. \textbf{ANLI R1} represents the data produced in the first round of this.
    \item \textbf{ANLI R3} \citep{nie2020anli} represents the data produced in the third round of the aforementioned iterative procedure. The adversarial model is trained on data produced in previous rounds, so crowdworkers are incentivized to create distinct entailment relations to challenge the model, so ANLI R3 is distinct from ANLI R1.
\end{itemize}  
For commonsense reasoning, we use the following:
\begin{itemize}
    \item \textbf{CommonsenseQA (CSQA)} \citep{talmor2019commonsenseqa} is a commonsense question answering dataset with multiple choice questions that require some prior knowledge not provided in the question.
    \item \textbf{RACE} \citep{lai2017race} is a commonsense question answering dataset where each question is related to a provided text passage. It evaluates understanding and reasoning abilities.
    \item \textbf{WinoGrande} \citep{sakaguchi2019winogrande} is a commonsense question answering dataset with fill-in-the-blank statements and binary answer options.
\end{itemize}

\noindent \textbf{Evaluation.} To evaluate the performance of different calibration datasets, we first prune the dense LLM with certain calibration data and then evaluate the resulting sparse LLM on all the downstream tasks considered using few-shot prompting~\citep{brown2020language}. 


\noindent \subsection{Calibration Data Formulation} The pruning calibration data have 128 sequences of length 2048 tokens each, following prior work~\cite{frantar2023massive,sun2023simple,yin2023outlier}.

\noindent \textbf{Pre-training Data.}
For each pre-training dataset, we create each calibration data sample of length 2048 tokens by concatenating text segments from the dataset until it exceeds 2048 tokens and then selecting a segment of length 2048 from this.

\noindent \textbf{Downstream Data.} To provide a comprehensive evaluation of downstream data, we use the following three variants.

\begin{itemize}

    \item \textbf{Zero-Shot}.  We create each calibration data sample by selecting a random question from the dataset without the answer. We fill up the remaining context length with padding tokens.
    \item \textbf{In-Context Learning.} We create each calibration data sample by concatenating multiple randomly selected question-answer pairs to fill up the context length of 2048 tokens.
    \item \textbf{In-Context Learning w/ Chain-of-Thought.} We create each calibration data sample by concatenating randomly selected question-answer pairs, where the answer contains CoT rationale, to fill up the context length of 2048 tokens.
\end{itemize}


\section{Results}

In this section, we report the results of our experiments. Our primary goal is to explore how performance fluctuates when using various calibration data across different formats. We analyze overall performance trends across these differing setups.

\subsection{Pre-training Dataset as Calibration Data}

We evaluate pruning performance using calibration data derived from a range of pre-training datasets including C4, RedPajama, Oscar, and Pile. The results are detailed in Table \ref{tab:combined_pretraining}. Our analysis reveals that the average accuracy of Pile consistently outperforms the C4 dataset.
Using Wanda with target sparsity 0.5, calibration with the Pile dataset exhibits superior performance in terms of average accuracy across nine downstream tasks, surpassing other pre-training datasets in six out of nine tasks. Similarly, for SparseGPT pruning, the Pile dataset achieves the highest average accuracy, although the differences among the four pre-training datasets are small. 

\begin{table*}[h]
\centering
\resizebox{\textwidth}{!}{
    \begin{tabular}{l|c|ccc|ccc}
    \toprule
    \textbf{Evaluation task} & \textbf{Dense Model} & \multicolumn{6}{c}{\textbf{Wanda \emph{w.} Calibration Data}} \\
    & & \multicolumn{1}{c}{GSM8K (Zero-shot)} & \multicolumn{1}{c}{GSM8K (ICL)} & \multicolumn{1}{c}{GSM8K (ICL w/ CoT)} & \multicolumn{1}{c}{e-SNLI (Zero-shot)} & \multicolumn{1}{c}{e-SNLI (ICL)} & \multicolumn{1}{c}{e-SNLI (ICL w/ CoT)} \\
    \midrule
    GSM8K & 0.0576 & 0.0205 & 0.0425 & \textbf{0.0432} & 0.0303 & \textbf{0.0432} & 0.0379 \\
    SVAMP & 0.3867 & 0.0233 & 0.2867 & \textbf{0.3067} & 0.1233 & 0.2100 & \textbf{0.2133} \\
    MAWPS & 0.4462 & 0.0058 & 0.3442 & \textbf{0.3519} & 0.0635 & \textbf{0.2635} & 0.2404 \\
    \midrule
    e-SNLI & 0.6050 & 0.3292 & \textbf{0.5438} & 0.5080 & 0.3428 & \textbf{0.5541} & 0.5517 \\
    ANLI R1 & 0.3900 & 0.2920 & \textbf{0.3180} & 0.3050 & 0.3340 & \textbf{0.3350} & 0.3330 \\
    ANLI R3 & 0.4192 & 0.2417 & \textbf{0.3567} & 0.3108 & 0.3350 & 0.3450 & \textbf{0.3717} \\
    \midrule
    CSQA & 0.6208 & 0.2138 & \textbf{0.5381} & 0.5184 & 0.4087 & 0.5127 & \textbf{0.5201} \\
    RACE & 0.6501 & 0.2067 & \textbf{0.4793} & 0.4698 & 0.3522 & 0.4653 & \textbf{0.4710} \\
    WinoGrande & 0.5122 & 0.5114 & 0.5130 & \textbf{0.5154} & 0.5051 & \textbf{0.5091} & 0.5075 \\
    \midrule
    Average & 0.4542 & 0.2049 & \textbf{0.3803} & 0.3699 & 0.2772 & 0.3598 & \textbf{0.3607} \\
    \bottomrule
        \end{tabular}}

\caption{Accuracy of Llama 2-Chat 7B model pruned with Wanda to 50\% unstructured sparsity using different formats of GSM8K and e-SNLI as calibration data. For each evaluation task, the best performance among the GSM8K calibration data variants and the best performance among the e-SNLI calibration data variants is bold.}
\label{tab:wanda_gsm8k_CoT}
\end{table*}

\begin{table*}[htbp]
    \centering
    \resizebox{\textwidth}{!}{%
    \begin{tabular}{l|c|c|c|ccc|ccc|ccc}
        \toprule
        \textbf{Model} & \textbf{Evaluation} & \textbf{Dense} & \multicolumn{10}{c}{\textbf{Wanda \emph{w.} Calibration Data}} \\
        \cmidrule{4-13}

       & & & \multicolumn{1}{c}{\textbf{PD}} & \multicolumn{3}{c}{\textbf{Arithmetic Reasoning}} & \multicolumn{3}{c}{\textbf{NLI}} & \multicolumn{3}{c}{\textbf{Commonsense Reasoning}} \\
        & & & \textbf{Pile} & \textbf{GSM8K} & \textbf{SVAMP} & \textbf{MAWPS} & \textbf{e-SNLI} & \textbf{ANLI R1} & \textbf{ANLI R3} & \textbf{CSQA} & \textbf{RACE} & \textbf{WinoGrande} \\
        \midrule
        \multirow{9}{*}{\rotatebox[origin=c]{90}{Llama 2-Chat 7B}} & GSM8K & 0.0576 & 0.0404 & 0.0425 & 0.0425 & \textbf{0.0462} & 0.0432 & 0.0417 & 0.0455 & 0.0417 & 0.0409 & 0.0432 \\
        & SVAMP & 0.3867 & \textbf{0.2878} & 0.2867 & 0.2833 & 0.2733 & 0.2100 & 0.2633 & 0.2667 & 0.2233 & 0.2667 & 0.2600 \\
        & MAWPS & 0.4462 & \textbf{0.3635} & 0.3442 & 0.3365 & 0.3346 & 0.2635 & 0.3038 & 0.3038 & 0.2654 & 0.3231 & 0.2731 \\
        \cmidrule{2-13}
        & e-SNLI & 0.6050 & 0.5376 & 0.5438 & 0.5711 & 0.5436 & 0.5541 & 0.5345 & 0.5441 & 0.5768 & 0.5317 & \textbf{0.5955} \\
        & ANLI R1 & 0.3900 & 0.3420 & 0.3180 & 0.3440 & 0.313 & 0.3350 & 0.3500 & 0.3490 & 0.3360 & 0.3370 & \textbf{0.3520} \\
        & ANLI R3 & 0.4192 & 0.3597 & 0.3567 & \textbf{0.3875} & 0.3700 & 0.3450 & 0.3700 & 0.3575 & 0.3633 & 0.3642 & 0.3792 \\
        \cmidrule{2-13}
        & CSQA & 0.6208 & 0.5239 & 0.5381 & 0.5233 & 0.5045 & 0.5127 & 0.5045 & 0.5364 & \textbf{0.5479} & 0.5373 & 0.5070 \\
        & RACE & 0.6501 & 0.4692 & \textbf{0.4793} & \textbf{0.4793} & 0.4726 & 0.4653 & 0.4341 & 0.4645 & 0.4706 & 0.4625 & 0.4422 \\
        & WinoGrande & 0.5122 & 0.5125 & 0.5130 & 0.5162 & 0.5114 & 0.5091 & \textbf{0.5257} & 0.5241 & 0.5107 & 0.5162 & 0.5209 \\
        \cmidrule{2-13}
        & \textbf{Average} & 0.4542 & 0.3819 & 0.3803 & \textbf{0.3871} & 0.3744 & 0.3598 & 0.3697 & 0.3768 & 0.3706 & 0.3755 & 0.3748 \\
        \toprule
        \multirow{9}{*}{\rotatebox[origin=c]{90}{LLaMA 7B}} & GSM8K & 0.0447 & 0.0409 & \textbf{0.0462} & 0.0440 & 0.0417 & 0.0394 & 0.0394 & 0.0417 & \textbf{0.0462} & 0.0387 & 0.0447 \\
        & SVAMP & 0.3267 & \textbf{0.2733} & 0.1533 & 0.2533 & 0.1900 & 0.1833 & 0.0867 & 0.1067 & 0.0733 & 0.0967 & 0.0800 \\
        & MAWPS & 0.3596 & 0.3173 & 0.3327 & \textbf{0.3577} & 0.3096 & 0.1615 & 0.2942 & 0.2846 & 0.1615 & 0.2808 & 0.2385 \\
        \cmidrule{2-13}
        & e-SNLI & 0.5556 & 0.3284 & 0.3433 & \textbf{0.3767} & 0.3678 & 0.3653 & 0.3430 & 0.3411 & 0.3304 & 0.3306 & 0.3291 \\
        & ANLI R1 & 0.3800 & 0.3210 & \textbf{0.4000} & \textbf{0.4000} & 0.3700 & 0.3340 & 0.2600 & 0.3100 & 0.3800 & 0.2600 & 0.3900 \\
        & ANLI R3 & 0.3167 & 0.3625 & 0.3833 & 0.3833 & 0.3750 & 0.3317 & 0.3583 & \textbf{0.4167} & 0.3417 & 0.3667 & 0.3917 \\
        \cmidrule{2-13}
        & CSQA & 0.3948 & 0.2613 & \textbf{0.2907} & 0.2793 & 0.2523 & 0.1974 & 0.2604 & 0.2735 & 0.2629 & 0.2752 & 0.2883 \\
        & RACE & 0.3134 & 0.2758 & 0.2972 & 0.2748 & 0.2525 & 0.2839 & 0.2657 & \textbf{0.3103} & 0.2698 & 0.2880 & 0.2748 \\
        & WinoGrande & 0.5130 & 0.4964 & 0.5154 & 0.5067 & \textbf{0.5264} & 0.5138 & 0.5162 & 0.5036 & 0.5043 & 0.5178 & 0.5225 \\
        \cmidrule{2-13}
        & \textbf{Average} & 0.3561 & 0.2974 & 0.3069 & \textbf{0.3195} & 0.2984 & 0.2678 & 0.2693 & 0.2876 & 0.2633 & 0.2727 & 0.2844 \\
        \bottomrule
    \end{tabular}
    }
    \caption{Accuracy of Llama 2-Chat 7B model and LLaMA 7B model pruned with Wanda to 50\% sparsity using various downstream datasets with ICL format. PD denotes pre-training data. The best performance on each evaluation task among sparse models is bold.} 
    \label{tab:wanda_icl}
\end{table*}

Notably, when compared with the commonly used C4 dataset, our analysis reveals that RedPajama achieves comparable performance, and Pile demonstrates an improvement, outperforming C4 in Wanda pruning across a majority of downstream tasks. Specifically, using the Llama 2-Chat 7b model, Pile leads C4 in seven out of nine tasks when using Wanda. Although when using SparseGPT, Pile outperforms C4 in only four out of nine tasks, Pile still has higher average accuracy across nine tasks. In Table \ref{tab:combined_pretraining_70}, when we target 70\% sparsity, we can clearly see that RedPajama and Pile achieve significantly higher average accuracy than C4. 
These findings underscore that C4 is not the optimal choice of calibration data for LLM pruning. Pile consistently serves as better calibration data in LLM pruning.



\subsection{Downstream Dataset as Calibration Data}

While using pre-training datasets for pruning may preserve acquired knowledge, it is crucial to empirically validate this strategy and determine if alternative downstream datasets might yield superior results for pruning LLMs. To this end, we utilized downstream datasets both as calibration data for pruning and as benchmarks for evaluation.




We compare three formats of downstream data: Zero-Shot, ICL and ICL w/ CoT. We systematically assessed the pruning performance across various downstream tasks using different calibration data formats: single GSM8K question (Zero-Shot), concatenated GSM8K question-answer pairs (ICL), and concatenated GSM8K question-answer pairs with Chain of Thought (ICL w/ CoT). Our findings, detailed in Table \ref{tab:wanda_gsm8k_CoT}, reveal that ICL consistently enhances performance across all data categories compared to the baseline zero-shot approach, achieving an average accuracy improvement of 0.1754. We also observed that GSM8K (ICL w/ CoT) calibration data outperforms GSM8K (ICL) data in Arithmetic Reasoning tasks. An explanation for this could be that the step-by-step reasoning in CoT calibration data helps guide the pruning to better preserve the model weights for arithmetic reasoning. However, GSM8K (ICL) surpasses GSM8K (ICL w/ CoT) in average performance across a broader set of downstream tasks as GSM8K (ICL) outperforms GSM8K (ICL w/ CoT) for tasks outside of arithmetic reasoning. This may be because the step-by-step reasoning in CoT introduces biases that are detrimental when the sparse model is used outside of the domain of the calibration data.

We also compare the pruning performance of e-SNLI (Zero-Shot), e-SNLI (ICL) and e-SNLI (ICL w/ CoT) in Table \ref{tab:wanda_gsm8k_CoT}. We find that ICL again enhances performance compared to the baseline zero-shot format, with an average accuracy improvement of 0.0826. We also find that, compared to the ICL format, including CoT in the calibration data only improves performance on ANLI R3 among the three NLI evaluation tasks. For the other categories of evaluation tasks, we find that e-SNLI (ICL) and e-SNLI (ICL w/ CoT) have similar pruning performance, and the former is better for some tasks and the latter is better for others.


\begin{table*}[h]
\centering
\resizebox{0.85\textwidth}{!}{
\begin{tabular}{l|c|c|ccccc}
\toprule
\textbf{Evaluation task} & \textbf{Dense Model} & \multicolumn{5}{c}{\textbf{Sparse Model}} \\
& & Pile & \multicolumn{4}{c}{GSM8K (ICL w/ CoT)} \\
& & & \multicolumn{1}{c}{Default (any \# of steps of CoT)} & \multicolumn{1}{c}{3-Step CoT} & \multicolumn{1}{c}{4-Step CoT} & \multicolumn{1}{c}{5-Step CoT} \\
\midrule
GSM8K & 0.0576 & 0.0404 & \textbf{0.0432} & 0.0402 & 0.0409 & 0.0387 \\
SVAMP & 0.3867 & 0.2878 & 0.3067 & 0.3100 & \textbf{0.3133} & 0.3033 \\
MAWPS & 0.4462 & 0.3635 & 0.3519 & 0.3558 & 0.3673 & \textbf{0.3808} \\
\bottomrule
\end{tabular}}

\caption{Accuracy of Llama 2-Chat 7B model pruned with Wanda to 50\% sparsity using different numbers of steps of CoT in the calibration data. For instance, GSM8K (ICL w/ $x$-step CoT) indicates the calibration data consists of concatenations of several question-answer pairs where each answer has exactly $x$ steps of reasoning. The default configuration of GSM8K (ICL w/ CoT) has no restriction on the number of steps of CoT.}
\label{tab:wanda_CoT_steps}
\end{table*}

\begin{table*}[h]
    \centering
    \resizebox{0.85\textwidth}{!}{
    \begin{tabular}{l|c|c|c|c}
        \toprule
         Evaluation task & \textbf{Dense Model} & \textbf{ Calibration Data } & \bf \# In-Context Q-A Pairs & \textbf{Sparse Model} \\
        \midrule
        GSM8K        & 0.0576 & C4 & - & 0.0455 \\
        GSM8K & 0.0576 &  Pile & - & 0.0404 \\
                    \midrule
        GSM8K &  0.0576  & GSM8K  & 5 & 0.0288 \\
        GSM8K &  0.0576 &  GSM8K & 10 &  0.0440 \\
         GSM8K & 0.0576 &  GSM8K & 15 & 0.0455 \\
        GSM8K &  0.0576 &  GSM8K  & 20 & 0.0417 \\
        GSM8K &  0.0576 &  GSM8K & 25 & \textbf{0.0470} \\
        GSM8K & 0.0576 &  GSM8K & Fill Q-A pairs to sequence length (2048 tokens) & 0.0425 \\

        \bottomrule
    \end{tabular}}
\caption{Accuracy of Llama 2-Chat 7B model pruned with Wanda to 50\% unstructured sparsity using GSM8K with different calibration data lengths and pre-training data.}
\label{tab:wanda_number_incontext_results}
\end{table*}

\begin{table}[h]
\centering
    \resizebox{\columnwidth}{!}{
    \begin{tabular}{l|c|c|c|c}
    \toprule
    \textbf{Evaluation task} & \textbf{Dense Model} & \multicolumn{3}{c}{\textbf{Sparse Model}} \\
    & & \multicolumn{1}{c}{Pile} & \multicolumn{1}{c}{ellipses} &
    \multicolumn{1}{c}{random alphanumeric } \\
    \midrule
    GSM8K & 0.0576 & \textbf{0.0404} & 0.0273 & 0.0402 \\
    SVAMP & 0.3867 & \textbf{0.2878} & 0.0576 & 0.1433 \\
    MAWPS & 0.4462 & \textbf{0.3635} & 0.0096 & 0.1462 \\
    \midrule
    e-SNLI & 0.6050 & \textbf{0.5376} & 0.3295 & 0.3679 \\
    ANLI R1 & 0.3900 & \textbf{0.3420} & 0.3100 & 0.3250 \\
    ANLI R3 & 0.4192 & \textbf{0.3597} & 0.3300 & 0.3275 \\
    \midrule
    CSQA & 0.6208 & \textbf{0.5239} & 0.1925 & 0.3170 \\
    RACE & 0.6501 & \textbf{0.4692} & 0.2631 & 0.3293 \\
    WinoGrande & 0.5122 & \textbf{0.5125} & 0.4972 & 0.5043 \\
    \midrule
    Average & 0.4542 &  \textbf{0.3819} & 0.2241 & 0.2779 \\
    \bottomrule
        \end{tabular}}
\caption{Accuracy of Llama 2-Chat 7B model pruned with Wanda to 50\% unstructured sparsity using Pile, ellipses, and random alphanumeric characters.}
\vspace{1em}
\label{tab:wanda_ellipses}
\end{table}

\begin{table}[h]
    \centering
\resizebox{\columnwidth}{!}{
    \begin{tabular}{l|c|c|c}
        \toprule
        \textbf{Evaluation task} & \textbf{Dense Model} & \textbf{Pruning Input Length} & \textbf{Sparse Model} \\
        \midrule
        WikiText & \multirow{5}{*}{6.94} & 128 & 29.22 \\
        WikiText & & 256 & 15.72 \\
        WikiText & & 512 & 11.82 \\
        WikiText & & 1024 & 9.27 \\
        WikiText & & 2048 & \textbf{8.48} \\
        \bottomrule
    \end{tabular}}

\caption{Perplexity of Llama 2-Chat 7B model on WikiText pruned with Wanda to 50\% unstructured sparsity using different input lengths of C4.}
\label{tab:wanda_seqlen}
\end{table}

\subsection{Winning Dataset?}

We evaluated the performance of ICL tasks against the top-performing pre-training dataset, Pile, with both the Llama 2-Chat 7B and LLaMA 7B models and have presented our findings in Table \ref{tab:wanda_icl}. Specifically, using the Llama 2-Chat 7B model, in the Arithmetic Reasoning category, Pile led in two out of three tasks. For NLI and Commonsense Reasoning tasks, the best calibration datasets come from the downstream dataset and from different task categories.  Upon reviewing average performance across all tasks, we observed that Arithmetic Reasoning generally matched the performance of the best pre-training dataset, Pile. Notably, SVAMP emerged as the most effective dataset overall, outperforming Pile with an average accuracy margin of 0.52\% with the Llama 2-Chat 7B model and with an average accuracy margin of 2.21\% with the Llama 7B model. Consequently, SVAMP has been designated as the winning dataset.

Additionally, an intriguing observation from our study was that the optimal calibration data for each downstream task did not necessarily coincide with the data from the corresponding task itself. This suggests that calibration data efficacy may not be task-specific and invites further exploration into the dynamics of calibration data across varied contexts.

\section{Further Analysis}

\textbf{{Can we do better by including more steps in CoT?}}  In our previous construction of the calibration data, we selected question-answer pairs with no restriction on the number of steps in CoT in the answer. This inspires a follow-up question: does the number of steps of CoT rationale in the calibration data affect the sparse LLM's performance?  We investigated this by constructing calibration data by concatenating multiple question-answer pairs, where each answer rationale contains exactly $x$ steps. Since 1-step or 2-step CoT data was scarce, we performed this for $x=\{3, 4, 5\}$ as seen in Table \ref{tab:wanda_CoT_steps}. We find no clear relationship between the number of steps of CoT in calibration data and the performance of the sparse LLM. However, we note that it is possible to produce a better sparse LLM for a given task by restricting the calibration data to a specific number of steps, which may vary based on the evaluation task.

\noindent \textbf{{Does more Q-A pairs in ICL calibration data lead to a better sparse model?}}  To investigate this, we evaluated the pruning performance when calibration data contains 5, 10, 15, 20, and 25 Q-A pairs, filling the rest of the context window with padding tokens. Our default ICL calibration data fills the context window with Q-A pairs until it reaches length 2048 tokens, which in practice can be anywhere from 25 to 30 Q-A pairs. We compare the pruning performance of all of these calibration data formats in Table \ref{tab:wanda_number_incontext_results}. The results confirm our conjecture that an increase in in-context examples in the pruning calibration data generally correlates with enhanced performance of the sparse model.

\noindent \textbf{{How does input length affect the pruning performance?}}  In our main experiments, the calibration data for pruning consisted of 128 sequences, each 2048 tokens in length. It is crucial to investigate whether this specific token length is necessary for effective pruning. To address this question, we used the C4 dataset for calibration and systematically varied the calibration data lengths between 256, 512, 1024, and 2048 tokens. We then evaluated the perplexity of Llama 2-Chat 7B pruned to 50\% unstructured sparsity using Wanda. As detailed in Table \ref{tab:wanda_seqlen}, our findings confirm that increased input lengths correlate positively with improved model performance, aligning with our initial expectations.

\noindent \textbf{{Does input data for pruning have to be sensible?}} In our previous setup, calibration data for pruning is sourced from either pre-training datasets or task-specific downstream datasets. It is intriguing to compare this with the pruning performance of nonsense data calibration data, such as ellipses and random alphanumeric strings,
 in this context. Consequently, we substituted conventional calibration data with these unconventional types for pruning the Llama 2-Chat 7B model to 50\% unstructured sparsity using the Wanda pruning method. The performance outcomes are shown in Table \ref{tab:wanda_ellipses}. The results clearly show that the Pile dataset, which contains human-readable data, consistently outperforms both ellipses and random alphanumeric strings in nearly all cases except one scenario within the GSM8K task. Moreover, random alphanumeric data generally exhibited better performance compared to ellipses. Therefore we affirm 
 the importance of utilizing sensible calibration data for the effective pruning of LLMs.

\section{Conclusion}

This study critically examines the widely held belief that the C4 dataset is the optimal calibration choice for pruning LLMs. Through an extensive evaluation encompassing a variety of calibration data types—both pre-training and downstream datasets, our findings reveal that C4 does not hold universal superiority. Specifically, our analysis demonstrates that the pretraining dataset Pile consistently outperforms C4, while alternative downstream datasets, particularly those involving arithmetic reasoning tasks, yield comparable pruning outcomes.

Furthermore, our investigation into various downstream task formats has uncovered that In-Context Learning (ICL) offers significant benefits across all data categories. In-Context Learning w/ Chain-of-Thought (ICL w/ CoT) calibration is particularly effective in enhancing performance in arithmetic reasoning tasks. 
Our study advocates for a more nuanced selection and curation of calibration data, which could lead to more efficient and effective LLM pruning strategies, ultimately facilitating the deployment of more robust models in practical settings.

\section{Limitations}

Our study has several limitations. First, all experiments were conducted using the Llama 2-Chat 7B and LLaMA 7B models; we aim to expand our investigations to other LLM architectures and larger models. Second, our analysis was limited to the Wanda and SparseGPT pruning algorithms. Future work will explore a broader range of pruning methods. Third, we plan to evaluate the effects of combining multiple datasets on pruning performance. We  believe that our insights regarding calibration data will inspire further research within the community.

Another limitation of this work we aim to address in the future is that we have not rigorously investigated why Pile is better calibration data than C4 for LLM pruning. We conjecture the benefits come from that Pile is a more diverse dataset with higher quality of examples, which is designed such that models trained on it have improved downstream generalization capabilities, compared to the more noisy Common Crawl datasets like C4, as also pointed out in recent work in the context of LLM pretraining \cite{li2024datacomp}. As such, we believe Pile could provide more robust calibration data to guide the pruning of LLMs to optimize the performance of the sparse model on a variety of downstream tasks. We leave the investigation on the correlation between a dataset’s effectiveness for LLM pretraining and model pruning as a future direction to explore.

\section*{Acknowledgement}
S. Liu is funded by the Royal Society with the Newton International Fellowship.


\bibliography{custom}

\begin{thebibliography}{48}
\providecommand{\natexlab}[1]{#1}

\bibitem[{Agarwalla et~al.(2024)Agarwalla, Gupta, Marques, Pandit, Goin, Kurtic, Leong, Nguyen, Salem, Alistarh et~al.}]{agarwalla2024enabling}
Abhinav Agarwalla, Abhay Gupta, Alexandre Marques, Shubhra Pandit, Michael Goin, Eldar Kurtic, Kevin Leong, Tuan Nguyen, Mahmoud Salem, Dan Alistarh, et~al. 2024.
\newblock Enabling high-sparsity foundational llama models with efficient pretraining and deployment.
\newblock \emph{arXiv preprint arXiv:2405.03594}.

\bibitem[{Anil et~al.(2023)Anil, Dai, Firat, Johnson, Lepikhin, Passos, Shakeri, Taropa, Bailey, Chen et~al.}]{anil2023palm}
Rohan Anil, Andrew~M Dai, Orhan Firat, Melvin Johnson, Dmitry Lepikhin, Alexandre Passos, Siamak Shakeri, Emanuel Taropa, Paige Bailey, Zhifeng Chen, et~al. 2023.
\newblock Palm 2 technical report.
\newblock \emph{arXiv preprint arXiv:2305.10403}.

\bibitem[{Brown et~al.(2020)Brown, Mann, Ryder, Subbiah, Kaplan, Dhariwal, Neelakantan, Shyam, Sastry, Askell et~al.}]{brown2020language}
Tom Brown, Benjamin Mann, Nick Ryder, Melanie Subbiah, Jared~D Kaplan, Prafulla Dhariwal, Arvind Neelakantan, Pranav Shyam, Girish Sastry, Amanda Askell, et~al. 2020.
\newblock Language models are few-shot learners.
\newblock \emph{Advances in neural information processing systems}, 33:1877--1901.

\bibitem[{Camburu et~al.(2018)Camburu, Rockt\"{a}schel, Lukasiewicz, and Blunsom}]{camburu2018esnli}
Oana-Maria Camburu, Tim Rockt\"{a}schel, Thomas Lukasiewicz, and Phil Blunsom. 2018.
\newblock \href {https://proceedings.neurips.cc/paper_files/paper/2018/file/4c7a167bb329bd92580a99ce422d6fa6-Paper.pdf} {e-snli: Natural language inference with natural language explanations}.
\newblock In \emph{Advances in Neural Information Processing Systems}, volume~31. Curran Associates, Inc.

\bibitem[{Chen et~al.(2022)Chen, Fard, Lo, and Bryksin}]{chen2022transferability}
Fuxiang Chen, Fatemeh~H Fard, David Lo, and Timofey Bryksin. 2022.
\newblock On the transferability of pre-trained language models for low-resource programming languages.
\newblock In \emph{Proceedings of the 30th IEEE/ACM International Conference on Program Comprehension}, pages 401--412.

\bibitem[{Cobbe et~al.(2021)Cobbe, Kosaraju, Bavarian, Chen, Jun, Kaiser, Plappert, Tworek, Hilton, Nakano, Hesse, and Schulman}]{cobbe2021gsm8k}
Karl Cobbe, Vineet Kosaraju, Mohammad Bavarian, Mark Chen, Heewoo Jun, Lukasz Kaiser, Matthias Plappert, Jerry Tworek, Jacob Hilton, Reiichiro Nakano, Christopher Hesse, and John Schulman. 2021.
\newblock \href {https://arxiv.org/abs/2110.14168} {Training verifiers to solve math word problems}.
\newblock \emph{CoRR}, abs/2110.14168.

\bibitem[{Frankle and Carbin(2019)}]{frankle2018lottery}
Jonathan Frankle and Michael Carbin. 2019.
\newblock The lottery ticket hypothesis: Finding sparse, trainable neural networks.
\newblock In \emph{International Conference on Learning Representations (ICLR)}.

\bibitem[{Frantar and Alistarh(2023{\natexlab{a}})}]{frantar2023massive}
Elias Frantar and Dan Alistarh. 2023{\natexlab{a}}.
\newblock Massive language models can be accurately pruned in one-shot.
\newblock In \emph{International Conference on Machine Learning (ICML)}.

\bibitem[{Frantar and Alistarh(2023{\natexlab{b}})}]{frantar2023sparsegpt}
Elias Frantar and Dan Alistarh. 2023{\natexlab{b}}.
\newblock Sparsegpt: Massive language models can be accurately pruned in one-shot.
\newblock In \emph{International Conference on Machine Learning}, pages 10323--10337. PMLR.

\bibitem[{Gale et~al.(2019)Gale, Elsen, and Hooker}]{gale2019state}
Trevor Gale, Erich Elsen, and Sara Hooker. 2019.
\newblock The state of sparsity in deep neural networks.
\newblock \emph{arXiv preprint arXiv:1902.09574}.

\bibitem[{Gao et~al.(2020)Gao, Biderman, Black, Golding, Hoppe, Foster, Phang, He, Thite, Nabeshima et~al.}]{gao2020pile}
Leo Gao, Stella Biderman, Sid Black, Laurence Golding, Travis Hoppe, Charles Foster, Jason Phang, Horace He, Anish Thite, Noa Nabeshima, et~al. 2020.
\newblock The pile: An 800gb dataset of diverse text for language modeling.
\newblock \emph{arXiv preprint arXiv:2101.00027}.

\bibitem[{{Gemini Team} et~al.(2023){Gemini Team}, Anil, Borgeaud, Wu, Alayrac, Yu, Soricut, Schalkwyk, Dai, Hauth et~al.}]{team2023gemini}
{Gemini Team}, Rohan Anil, Sebastian Borgeaud, Yonghui Wu, Jean-Baptiste Alayrac, Jiahui Yu, Radu Soricut, Johan Schalkwyk, Andrew~M Dai, Anja Hauth, et~al. 2023.
\newblock Gemini: a family of highly capable multimodal models.
\newblock \emph{arXiv preprint arXiv:2312.11805}.

\bibitem[{Han et~al.(2015)Han, Pool, Tran, and Dally}]{han2015learning}
Song Han, Jeff Pool, John Tran, and William Dally. 2015.
\newblock Learning both weights and connections for efficient neural network.
\newblock In \emph{Advances in Neural Information Processing Systems (NeurIPS)}, pages 1135--1143.

\bibitem[{Hoang et~al.(2023)Hoang, Cho, Merth, Rastegari, and Wang}]{hoang2023dynamic}
Duc~NM Hoang, Minsik Cho, Thomas Merth, Mohammad Rastegari, and Zhangyang Wang. 2023.
\newblock (dynamic) prompting might be all you need to repair compressed llms.
\newblock \emph{arXiv preprint arXiv:2310.00867}.

\bibitem[{Jiang et~al.(2020)Jiang, Xu, Araki, and Neubig}]{jiang2020can}
Zhengbao Jiang, Frank~F Xu, Jun Araki, and Graham Neubig. 2020.
\newblock How can we know what language models know?
\newblock \emph{Transactions of the Association for Computational Linguistics}, 8:423--438.

\bibitem[{Koco{\'n} et~al.(2023)Koco{\'n}, Cichecki, Kaszyca, Kochanek, Szyd{\l}o, Baran, Bielaniewicz, Gruza, Janz, Kanclerz et~al.}]{kocon2023chatgpt}
Jan Koco{\'n}, Igor Cichecki, Oliwier Kaszyca, Mateusz Kochanek, Dominika Szyd{\l}o, Joanna Baran, Julita Bielaniewicz, Marcin Gruza, Arkadiusz Janz, Kamil Kanclerz, et~al. 2023.
\newblock Chatgpt: Jack of all trades, master of none.
\newblock \emph{Information Fusion}, 99:101861.

\bibitem[{Koncel-Kedziorski et~al.(2016)Koncel-Kedziorski, Roy, Amini, Kushman, and Hajishirzi}]{koncel2016mawps}
Rik Koncel-Kedziorski, Subhro Roy, Aida Amini, Nate Kushman, and Hannaneh Hajishirzi. 2016.
\newblock Mawps: A math word problem repository.
\newblock In \emph{Proceedings of the 2016 conference of the north american chapter of the association for computational linguistics: human language technologies}, pages 1152--1157.

\bibitem[{Lai et~al.(2017)Lai, Xie, Liu, Yang, and Hovy}]{lai2017race}
Guokun Lai, Qizhe Xie, Hanxiao Liu, Yiming Yang, and Eduard Hovy. 2017.
\newblock Race: Large-scale reading comprehension dataset from examinations.
\newblock \emph{arXiv preprint arXiv:1704.04683}.

\bibitem[{Lester et~al.(2021)Lester, Al-Rfou, and Constant}]{lester2021power}
Brian Lester, Rami Al-Rfou, and Noah Constant. 2021.
\newblock The power of scale for parameter-efficient prompt tuning.
\newblock \emph{arXiv preprint arXiv:2104.08691}.

\bibitem[{Li et~al.(2024)Li, Fang, Smyrnis, Ivgi, Jordan, Gadre, Bansal, Guha, Keh, Arora et~al.}]{li2024datacomp}
Jeffrey Li, Alex Fang, Georgios Smyrnis, Maor Ivgi, Matt Jordan, Samir Gadre, Hritik Bansal, Etash Guha, Sedrick Keh, Kushal Arora, et~al. 2024.
\newblock Datacomp-lm: In search of the next generation of training sets for language models.
\newblock \emph{arXiv preprint arXiv:2406.11794}.

\bibitem[{Liu et~al.(2022)Liu, Chen, Chen, Shen, Mocanu, Wang, and Pechenizkiy}]{liu2022unreasonable}
Shiwei Liu, Tianlong Chen, Xiaohan Chen, Li~Shen, Decebal~Constantin Mocanu, Zhangyang Wang, and Mykola Pechenizkiy. 2022.
\newblock The unreasonable effectiveness of random pruning: Return of the most naive baseline for sparse training.
\newblock \emph{arXiv preprint arXiv:2202.02643}.

\bibitem[{Mocanu et~al.(2018)Mocanu, Mocanu, Stone, Nguyen, Gibescu, and Liotta}]{mocanu2018scalable}
Decebal~Constantin Mocanu, Elena Mocanu, Peter Stone, Phuong~H Nguyen, Madeleine Gibescu, and Antonio Liotta. 2018.
\newblock Scalable training of artificial neural networks with adaptive sparse connectivity inspired by network science.
\newblock \emph{Nature Communications}, 9:1--12.

\bibitem[{Molchanov et~al.(2017)Molchanov, Tyree, Karras, Aila, and Kautz}]{molchanov2016pruning}
Pavlo Molchanov, Stephen Tyree, Tero Karras, Timo Aila, and Jan Kautz. 2017.
\newblock Pruning convolutional neural networks for resource efficient inference.
\newblock In \emph{International Conference on Learning Representations (ICLR)}.

\bibitem[{Mozer and Smolensky(1989)}]{mozer1989skeletonization}
Michael~C Mozer and Paul Smolensky. 1989.
\newblock Skeletonization: A technique for trimming the fat from a network via relevance assessment.
\newblock In \emph{Advances in Neural Information Processing Systems (NeurIPS)}, pages 107--115.

\bibitem[{Nie et~al.(2020)Nie, Williams, Dinan, Bansal, Weston, and Kiela}]{nie2020anli}
Yixin Nie, Adina Williams, Emily Dinan, Mohit Bansal, Jason Weston, and Douwe Kiela. 2020.
\newblock Adversarial {NLI}: A new benchmark for natural language understanding.
\newblock In \emph{Proceedings of the 58th Annual Meeting of the Association for Computational Linguistics}. Association for Computational Linguistics.

\bibitem[{Patel et~al.(2021)Patel, Bhattamishra, and Goyal}]{patel2021svamp}
Arkil Patel, Satwik Bhattamishra, and Navin Goyal. 2021.
\newblock \href {https://arxiv.org/abs/2103.07191} {Are {NLP} models really able to solve simple math word problems?}
\newblock \emph{CoRR}, abs/2103.07191.

\bibitem[{Radford et~al.(2021)Radford, Kim, Hallacy, Ramesh, Goh, Agarwal, Sastry, Askell, Mishkin, Clark et~al.}]{radford2021learning}
Alec Radford, Jong~Wook Kim, Chris Hallacy, Aditya Ramesh, Gabriel Goh, Sandhini Agarwal, Girish Sastry, Amanda Askell, Pamela Mishkin, Jack Clark, et~al. 2021.
\newblock Learning transferable visual models from natural language supervision.
\newblock In \emph{International conference on machine learning}, pages 8748--8763. PMLR.

\bibitem[{Raffel et~al.(2020)Raffel, Shazeer, Roberts, Lee, Narang, Matena, Zhou, Li, and Liu}]{raffel2020exploring}
Colin Raffel, Noam Shazeer, Adam Roberts, Katherine Lee, Sharan Narang, Michael Matena, Yanqi Zhou, Wei Li, and Peter~J Liu. 2020.
\newblock Exploring the limits of transfer learning with a unified text-to-text transformer.
\newblock \emph{The Journal of Machine Learning Research}, 21(1):5485--5551.

\bibitem[{Romera-Paredes et~al.(2024)Romera-Paredes, Barekatain, Novikov, Balog, Kumar, Dupont, Ruiz, Ellenberg, Wang, Fawzi et~al.}]{romera2024mathematical}
Bernardino Romera-Paredes, Mohammadamin Barekatain, Alexander Novikov, Matej Balog, M~Pawan Kumar, Emilien Dupont, Francisco~JR Ruiz, Jordan~S Ellenberg, Pengming Wang, Omar Fawzi, et~al. 2024.
\newblock Mathematical discoveries from program search with large language models.
\newblock \emph{Nature}, 625(7995):468--475.

\bibitem[{Sakaguchi et~al.(2019)Sakaguchi, Bras, Bhagavatula, and Choi}]{sakaguchi2019winogrande}
Keisuke Sakaguchi, Ronan~Le Bras, Chandra Bhagavatula, and Yejin Choi. 2019.
\newblock Winogrande: An adversarial winograd schema challenge at scale.
\newblock \emph{arXiv preprint arXiv:1907.10641}.

\bibitem[{Shi et~al.(2023)Shi, Chen, Misra, Scales, Dohan, Chi, Sch{\"a}rli, and Zhou}]{shi2023large}
Freda Shi, Xinyun Chen, Kanishka Misra, Nathan Scales, David Dohan, Ed~H Chi, Nathanael Sch{\"a}rli, and Denny Zhou. 2023.
\newblock Large language models can be easily distracted by irrelevant context.
\newblock In \emph{International Conference on Machine Learning}, pages 31210--31227. PMLR.

\bibitem[{Su{\'a}rez et~al.(2020)Su{\'a}rez, Romary, and Sagot}]{suarez2020monolingual}
Pedro Javier~Ortiz Su{\'a}rez, Laurent Romary, and Beno{\^\i}t Sagot. 2020.
\newblock A monolingual approach to contextualized word embeddings for mid-resource languages.
\newblock \emph{arXiv preprint arXiv:2006.06202}.

\bibitem[{Sun et~al.(2023)Sun, Liu, Bair, and Kolter}]{sun2023simple}
Mingjie Sun, Zhuang Liu, Anna Bair, and J~Zico Kolter. 2023.
\newblock A simple and effective pruning approach for large language models.
\newblock \emph{arXiv preprint arXiv:2306.11695}.

\bibitem[{Talmor et~al.(2019)Talmor, Herzig, Lourie, and Berant}]{talmor2019commonsenseqa}
Alon Talmor, Jonathan Herzig, Nicholas Lourie, and Jonathan Berant. 2019.
\newblock \href {https://doi.org/10.18653/v1/N19-1421} {{C}ommonsense{QA}: A question answering challenge targeting commonsense knowledge}.
\newblock In \emph{Proceedings of the 2019 Conference of the North {A}merican Chapter of the Association for Computational Linguistics: Human Language Technologies, Volume 1 (Long and Short Papers)}, pages 4149--4158, Minneapolis, Minnesota. Association for Computational Linguistics.

\bibitem[{{Together Computer}(2023)}]{together2023RedPajama}
{Together Computer}. 2023.
\newblock \href {https://github.com/togethercomputer/RedPajama-Data} {Redpajama: an open dataset for training large language models}.

\bibitem[{{Touvron} et~al.(2023){Touvron}, {Martin}, {Stone}, {Albert}, {Almahairi}, {Babaei}, {Bashlykov}, {Batra}, {Bhargava}, {Bhosale}, {Bikel}, {Blecher}, {Canton Ferrer}, {Chen}, {Cucurull}, {Esiobu}, {Fernandes}, {Fu}, {Fu}, {Fuller}, {Gao}, {Goswami}, {Goyal}, {Hartshorn}, {Hosseini}, {Hou}, {Inan}, {Kardas}, {Kerkez}, {Khabsa}, {Kloumann}, {Korenev}, {Singh Koura}, {Lachaux}, {Lavril}, {Lee}, {Liskovich}, {Lu}, {Mao}, {Martinet}, {Mihaylov}, {Mishra}, {Molybog}, {Nie}, {Poulton}, {Reizenstein}, {Rungta}, {Saladi}, {Schelten}, {Silva}, {Smith}, {Subramanian}, {Tan}, {Tang}, {Taylor}, {Williams}, {Kuan}, {Xu}, {Yan}, {Zarov}, {Zhang}, {Fan}, {Kambadur}, {Narang}, {Rodriguez}, {Stojnic}, {Edunov}, and {Scialom}}]{touvron2023llama2}
Hugo {Touvron}, Louis {Martin}, Kevin {Stone}, Peter {Albert}, Amjad {Almahairi}, Yasmine {Babaei}, Nikolay {Bashlykov}, Soumya {Batra}, Prajjwal {Bhargava}, Shruti {Bhosale}, Dan {Bikel}, Lukas {Blecher}, Cristian {Canton Ferrer}, Moya {Chen}, Guillem {Cucurull}, David {Esiobu}, Jude {Fernandes}, Jeremy {Fu}, Wenyin {Fu}, Brian {Fuller}, Cynthia {Gao}, Vedanuj {Goswami}, Naman {Goyal}, Anthony {Hartshorn}, Saghar {Hosseini}, Rui {Hou}, Hakan {Inan}, Marcin {Kardas}, Viktor {Kerkez}, Madian {Khabsa}, Isabel {Kloumann}, Artem {Korenev}, Punit {Singh Koura}, Marie-Anne {Lachaux}, Thibaut {Lavril}, Jenya {Lee}, Diana {Liskovich}, Yinghai {Lu}, Yuning {Mao}, Xavier {Martinet}, Todor {Mihaylov}, Pushkar {Mishra}, Igor {Molybog}, Yixin {Nie}, Andrew {Poulton}, Jeremy {Reizenstein}, Rashi {Rungta}, Kalyan {Saladi}, Alan {Schelten}, Ruan {Silva}, Eric~Michael {Smith}, Ranjan {Subramanian}, Xiaoqing~Ellen {Tan}, Binh {Tang}, Ross {Taylor}, Adina {Williams}, Jian~Xiang {Kuan}, Puxin {Xu}, Zheng {Yan}, Iliyan {Zarov},
  Yuchen {Zhang}, Angela {Fan}, Melanie {Kambadur}, Sharan {Narang}, Aurelien {Rodriguez}, Robert {Stojnic}, Sergey {Edunov}, and Thomas {Scialom}. 2023.
\newblock \href {https://doi.org/10.48550/arXiv.2307.09288} {{Llama 2: Open Foundation and Fine-Tuned Chat Models}}.
\newblock \emph{arXiv e-prints}, arXiv:2307.09288.

\bibitem[{Touvron et~al.(2023)Touvron, Martin, Stone, Albert, Almahairi, Babaei, Bashlykov, Batra, Bhargava, Bhosale et~al.}]{touvron2023llama}
Hugo Touvron, Louis Martin, Kevin Stone, Peter Albert, Amjad Almahairi, Yasmine Babaei, Nikolay Bashlykov, Soumya Batra, Prajjwal Bhargava, Shruti Bhosale, et~al. 2023.
\newblock Llama 2: Open foundation and fine-tuned chat models.
\newblock \emph{arXiv preprint arXiv:2307.09288}.

\bibitem[{Wang et~al.(2023)Wang, Qin, Bai, and Fu}]{wang2023state}
Huan Wang, Can Qin, Yue Bai, and Yun Fu. 2023.
\newblock Why is the state of neural network pruning so confusing? on the fairness, comparison setup, and trainability in network pruning.
\newblock \emph{arXiv preprint arXiv:2301.05219}.

\bibitem[{Wei et~al.(2022)Wei, Wang, Schuurmans, Bosma, Xia, Chi, Le, Zhou et~al.}]{wei2022chain}
Jason Wei, Xuezhi Wang, Dale Schuurmans, Maarten Bosma, Fei Xia, Ed~Chi, Quoc~V Le, Denny Zhou, et~al. 2022.
\newblock Chain-of-thought prompting elicits reasoning in large language models.
\newblock \emph{Advances in Neural Information Processing Systems (NeurIPs)}, 35:24824--24837.

\bibitem[{Williams and Aletras(2023)}]{williams2023does}
Miles Williams and Nikolaos Aletras. 2023.
\newblock How does calibration data affect the post-training pruning and quantization of large language models?
\newblock \emph{arXiv preprint arXiv:2311.09755}.

\bibitem[{Xu et~al.(2023)Xu, Liu, Chen, Tang, Wang, Zhou, Hu, and Shrivastava}]{xu2023compress}
Zhaozhuo Xu, Zirui Liu, Beidi Chen, Yuxin Tang, Jue Wang, Kaixiong Zhou, Xia Hu, and Anshumali Shrivastava. 2023.
\newblock Compress, then prompt: Improving accuracy-efficiency trade-off of llm inference with transferable prompt.
\newblock \emph{arXiv preprint arXiv:2305.11186}.

\bibitem[{Yin et~al.(2023{\natexlab{a}})Yin, Li, Fang, Shen, Huang, Wang, Menkovski, Ma, Pechenizkiy, and Liu}]{yin2023dynamic}
Lu~Yin, Gen Li, Meng Fang, Li~Shen, Tianjin Huang, Zhangyang Wang, Vlado Menkovski, Xiaolong Ma, Mykola Pechenizkiy, and Shiwei Liu. 2023{\natexlab{a}}.
\newblock Dynamic sparsity is channel-level sparsity learner.
\newblock \emph{arXiv preprint arXiv:2305.19454}.

\bibitem[{Yin et~al.(2023{\natexlab{b}})Yin, Wu, Zhang, Hsieh, Wang, Jia, Pechenizkiy, Liang, Wang, and Liu}]{yin2023outlier}
Lu~Yin, You Wu, Zhenyu Zhang, Cheng-Yu Hsieh, Yaqing Wang, Yiling Jia, Mykola Pechenizkiy, Yi~Liang, Zhangyang Wang, and Shiwei Liu. 2023{\natexlab{b}}.
\newblock Outlier weighed layerwise sparsity (owl): A missing secret sauce for pruning llms to high sparsity.
\newblock \emph{arXiv preprint arXiv:2310.05175}.

\bibitem[{Zhang et~al.(2023)Zhang, Zhao, Lin, Sun, Yao, Han, Tanner, Liu, and Ji}]{zhang2023dynamic}
Yuxin Zhang, Lirui Zhao, Mingbao Lin, Yunyun Sun, Yiwu Yao, Xingjia Han, Jared Tanner, Shiwei Liu, and Rongrong Ji. 2023.
\newblock Dynamic sparse no training: Training-free fine-tuning for sparse llms.
\newblock \emph{arXiv preprint arXiv:2310.08915}.

\bibitem[{Zhao et~al.(2021)Zhao, Wallace, Feng, Klein, and Singh}]{zhao2021calibrate}
Zihao Zhao, Eric Wallace, Shi Feng, Dan Klein, and Sameer Singh. 2021.
\newblock Calibrate before use: Improving few-shot performance of language models.
\newblock In \emph{International conference on machine learning}, pages 12697--12706. PMLR.

\bibitem[{Zhou et~al.(2022{\natexlab{a}})Zhou, Yang, Loy, and Liu}]{zhou2022conditional}
Kaiyang Zhou, Jingkang Yang, Chen~Change Loy, and Ziwei Liu. 2022{\natexlab{a}}.
\newblock Conditional prompt learning for vision-language models.
\newblock In \emph{Proceedings of the IEEE/CVF conference on computer vision and pattern recognition}, pages 16816--16825.

\bibitem[{Zhou et~al.(2022{\natexlab{b}})Zhou, Muresanu, Han, Paster, Pitis, Chan, and Ba}]{zhou2022large}
Yongchao Zhou, Andrei~Ioan Muresanu, Ziwen Han, Keiran Paster, Silviu Pitis, Harris Chan, and Jimmy Ba. 2022{\natexlab{b}}.
\newblock Large language models are human-level prompt engineers.
\newblock \emph{arXiv preprint arXiv:2211.01910}.

\bibitem[{Zimmer et~al.(2023)Zimmer, Andoni, Spiegel, and Pokutta}]{zimmer2023perp}
Max Zimmer, Megi Andoni, Christoph Spiegel, and Sebastian Pokutta. 2023.
\newblock Perp: Rethinking the prune-retrain paradigm in the era of llms.
\newblock \emph{arXiv preprint arXiv:2312.15230}.

\end{thebibliography}

\clearpage
\newpage

\end{document}